\documentclass[conference]{IEEEtran}
\IEEEoverridecommandlockouts
\usepackage{cite}
\usepackage{amsmath,amssymb,amsfonts}
\usepackage{algorithmic}
\usepackage{graphicx}
\usepackage{textcomp}
\usepackage{xcolor}
\usepackage{multirow}
\usepackage{siunitx}
\def\BibTeX{{\rm B\kern-.05em{\sc i\kern-.025em b}\kern-.08em
    T\kern-.1667em\lower.7ex\hbox{E}\kern-.125emX}}
    
\usepackage{fancyhdr}
\begin{document}

\title{Attentive Cross-modal Connections for Deep Multimodal Wearable-based Emotion Recognition\\

% \thanks{We would like to acknowledge the Innovation for Defence Excellence and Security (IDEaS) program for funding this project.}
}

% \author{\IEEEauthorblockN{1\textsuperscript{st} Anubhav Bhatti}
% \IEEEauthorblockA{\textit{Department of Electrical and Computer Engineering} \\
% \textit{Queen's University}\\
% Kingston, Canada\\
% anubhav.bhatti@queensu.ca}
% \and
% \IEEEauthorblockN{2\textsuperscript{nd} Behnam Behinaein}
% \IEEEauthorblockA{\textit{Department of Electrical and Computer Engineering} \\
% \textit{Queen's University}\\
% Kingston, Canada\\
% hbb1@queensu.ca}
% \and
% \IEEEauthorblockN{3\textsuperscript{rd} Paul Hungler}
% \IEEEauthorblockA{\textit{Ingenuity Labs Research Institute} \\
% \textit{Queen's University}\\
% Kingston, Canada\\
% paul.hungler@queensu.ca}
% \and
% \IEEEauthorblockN{4\textsuperscript{th} Ali Etemad}
% \IEEEauthorblockA{\textit{Department of Electrical and Computer Engineering} \\
% \textit{Queen's University} \\
% Kingston, Canada\\
% ali.etemad@queensu.ca}
% }

\author{\IEEEauthorblockN{Anubhav Bhatti$^{1, 2}$, Behnam Behinaein$^{1, 2}$, Dirk Rodenburg$^{2}$, Paul Hungler$^{2}$, Ali Etemad$^{1, 2}$}
\IEEEauthorblockA{\textit{$^{1}$Department of Electrical and Computer Engineering, $^{2}$Ingenuity Labs Research Institute} \\
\textit{Queen's University}\\
Kingston, Canada\\
(anubhav.bhatti, hbb1, djr08, paul.hungler, ali.etemad)@queensu.ca}
}

\maketitle
\thispagestyle{fancy}

\begin{abstract}
Classification of human emotions can play an essential role in the design and improvement of human-machine systems. While individual biological signals such as Electrocardiogram (ECG) and Electrodermal Activity (EDA) have been widely used for emotion recognition with machine learning methods, multimodal approaches generally fuse extracted features or final classification/regression results to boost performance. To enhance multimodal learning, we present a novel attentive cross-modal connection to share information between convolutional neural networks responsible for learning individual modalities. Specifically, these connections improve emotion classification by sharing intermediate representations among EDA and ECG and apply attention weights to the shared information, thus learning more effective multimodal embeddings. We perform experiments on the WESAD dataset to identify the best configuration of the proposed method for emotion classification. Our experiments show that the proposed approach is capable of learning strong multimodal representations and outperforms a number of baselines methods.
\end{abstract}


\section{Introduction}
\label{sec:intro}
\textit{Affective computing} is an evolving field that focuses on the development of systems and devices to recognize, interpret, process, and simulate human emotions \cite{picard2000affective, tao2005affective}. Previous studies have shown that signals such as Electrocardiogram (ECG) and Electrodermal Activity (EDA), also called Galvanic Skin Response, have the potential to detect affective states such as stress levels in humans \cite{sioni2015stress, hsieh2019feature, setz2009discriminating}. Generally, different deep learning models can be used to extract features from each modality to be used in classifiers for predicting the emotion class \cite{hwang2018deep, sarkar2019classification, sarkar2020self, sarkar2020self_icassp}. On the other hand, different modalities often contain rich complementary or overlapping information that needs to be considered to boost performance through \textit{fusion}-based representation learning strategies. Integrating multiple modalities, their learned features, and/or the intermediate decisions of machine learning models responsible for processing them for a discriminative task is referred to as \emph{multimodal fusion}.

\textit{Feature-level fusion}, also referred to as \textit{early fusion}, and \textit{decision-level fusion}, also called \textit{late fusion}, are two popular approaches for combining different modalities \cite{lin2019explainable, plarre2011continuous, siddharth2019utilizing, bota2020emotion}. In the former strategy, representations extracted by the feature extractors are combined and then fed to a classifier for the classification task, while in the latter, decisions obtained based on the representations of individual modalities are combined to reach a final decision.

Recent works \cite{velivckovic2016xcnn,cangea2019xflow} have shown that \textit{cross-connections} between different hidden layers of neural networks responsible for learning different modalities may result in an effective exchange of learned representations for exploiting complementary or avoiding redundant information. This, in turn, may result in better performance. However, those studies have focused on directly connecting different modality streams without performing any additional processing on the exchanged information. Moreover, exhaustive experiments have not been presented to show the impact of different possible configurations of such cross-modal connections, for example, their location within a deep multimodal pipeline, or the source and the target modality of the shared information.

In this paper, we expand on this notion and propose a novel attentive cross-modal connection (henceforth, referred to as \textit{AttX connections}) to learn strong shared representations for multimodal (ECG-EDA) affective computing. The cross-connections between the networks responsible for learning each modality use an attention mechanism to regulate the fusion of intermediate representations by learning the importance of each modality while training. We introduce three different variations of these connections for sharing information from ECG to EDA (Type I), EDA to ECG (Type II), and simultaneously sharing information between both (Type III).

Our contributions in this paper can be summarized as follows.
(\textbf{1}) We extend the notion of cross-modal connections by introducing an attention module to regulate the exchange of information between modalities. We also introduce three variations of the proposed connections to control the flow of information better.
(\textbf{2}) We apply our proposed method to multimodal ECG-EDA emotion recognition and perform extensive experiments on the publicly available WESAD dataset to show the flexibility and impact of our method in obtaining better or competitive performance to the state-of-the-art.
(\textbf{3}) Our experiments show that bi-directional sharing of information between ECG and EDA results in stronger and more effective representations instead of sharing ECG information with EDA alone or vice-versa. Moreover, our study shows that adding the proposed mechanism in multiple different locations within a multimodal pipeline is advantageous vs. restricting information sharing/fusion to a single point.

\section{Proposed Method}
\label{sec:pagestyle}

\subsection{Problem Setup}
Let's assume we have two deep neural networks $H_{ecg}$ and $H_{eda}$, trained to learn and classify representations from ECG and EDA modalities, respectively. Existing fusion techniques mostly rely on fusing (concatenating) the representations learned by $H_{ecg}$ and $H_{eda}$ prior to the fully connected (FC) layers intended for the final classification, or fusing the outputs of $H_{ecg}$ and $H_{eda}$ by [weighted] averaging or voting. While cross-modal connections, which have been proposed in the past \cite{velivckovic2016xcnn,cangea2019xflow} can be used for fusing $H_{ecg}$ and $H_{eda}$, such methods fail to regulate the flow of information between the two modalities, thus fail to adaptively control fused information where necessary.

\begin{figure}[!t]
  \centering
   \includegraphics[width=1\columnwidth, height=1\linewidth, keepaspectratio]{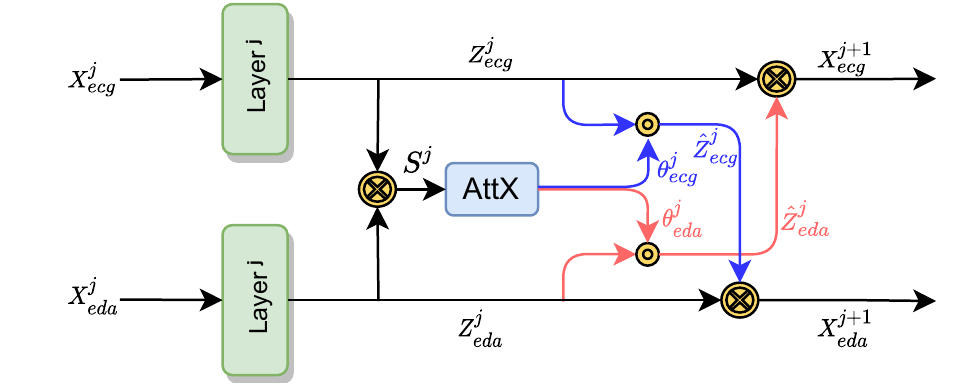} % images/attx_types.pdf
  \caption{The three configurations for the proposed AttX is presented: Type I: ECG to EDA (Blue), Type II: EDA to ECG (Red), and Type III: ECG to EDA and EDA to ECG (Blue and Red together). The symbol $\bigotimes$ represents concatenation, and $\bigodot$ represents element-wise multiplication.}
  \label{fig:attx_types}
\end{figure}

\subsection{Attentive Cross-modal Connections (AttX)}
\label{sec:attention}
In this section, details of attentive cross-modal connections are discussed. As shown in Figure \ref{fig:attx_types}, we denote the inputs to the convolutional block $j$ of ECG and EDA networks by $X^j_{ecg}$ and $X^j_{eda}$, respectively. Accordingly, the input ECG and EDA to the network (first block) are $X^1_{ecg}$ and $X^1_{eda}$, respectively. The learned representation obtained from feeding $X^1_{ecg}$ to $H_{ecg}$ and $X^1_{eda}$ to $H_{eda}$, after convolutional block $j$ of the network can be denoted by $Z^{j}_{ecg}$ and $Z^{j}_{eda}$, respectively. To obtain rich multimodal representations through intermediate fusions of $Z^{j}_{ecg}$ and $Z^{j}_{eda}$ while allowing for flexible and adaptive regulation of the shared information, we propose attentive cross-connections (AttX) between $H_{ecg}$ and $H_{eda}$. Attention is a mechanism that focuses on the most salient parts of the input or features by emphasizing (weights) on a subset of the feature set. Such mechanisms have been widely employed in natural language processing \cite{vaswani2017attention}, computer vision \cite{sun2020survey}, and bio-signal analysis \cite{zhang2020classification}. Here, we intend to utilize this mechanism for regulating the flow of information through intermediate cross-modal connections.

\begin{figure*}[!t]
  \centering
  \includegraphics[width=0.85\linewidth, keepaspectratio]{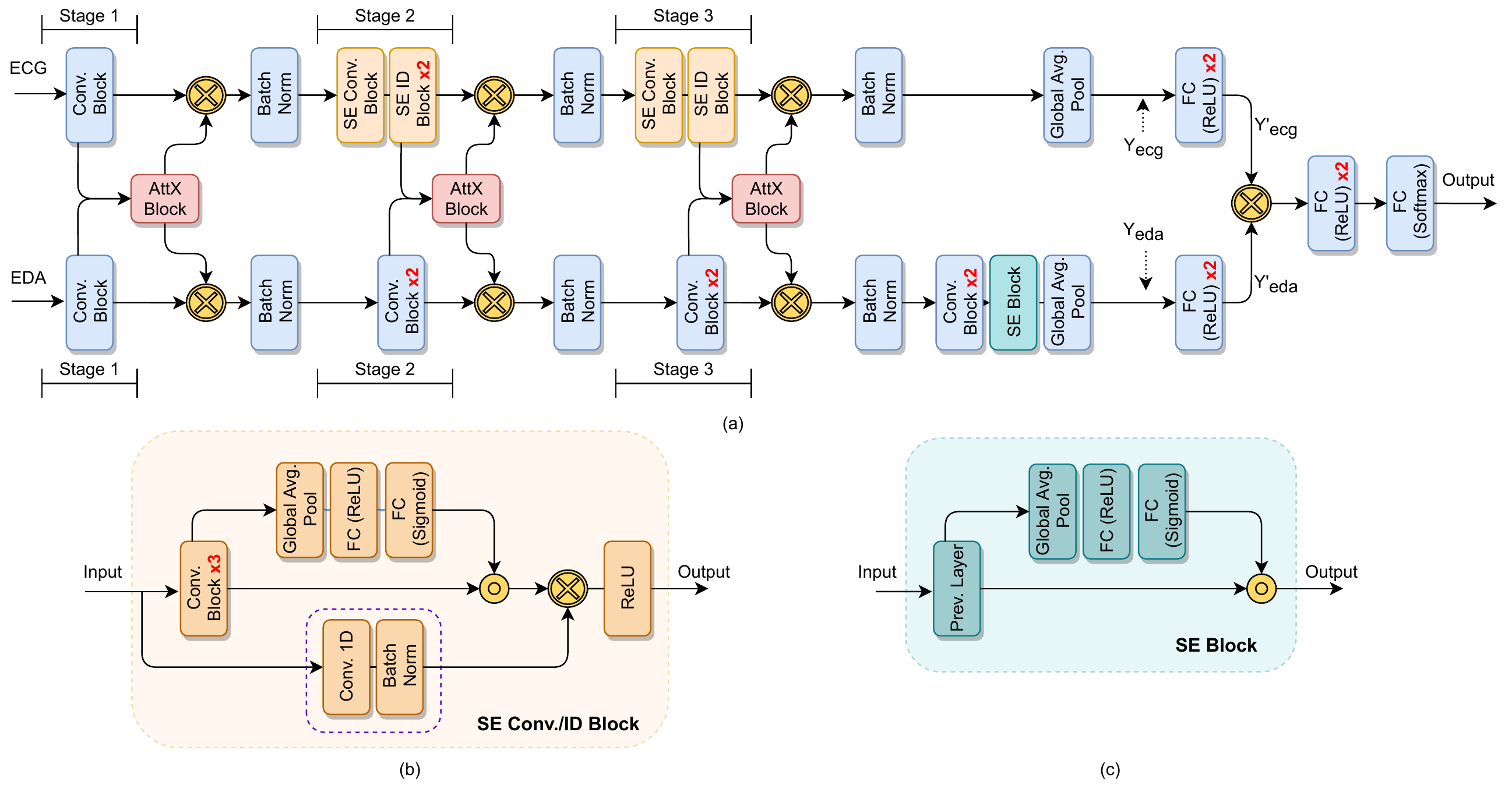} %arch_2_4.pdf
  \caption{Our proposed multimodal pipeline along with AttX connections stress classification (a). The symbol $\bigoplus$ represents addition, $\bigotimes$ represents concatenation, and $\bigodot$ represents element-wise multiplication. The SE Conv./ID Block is presented (b), where removing the purple block results in the SE ID Block. The SE Block for EDA is presented in (c).}
  \label{fig:networkarch}
\end{figure*}

% ***
To do so, we first concatenate $Z^{j}_{ecg}$ and $Z^{j}_{eda}$ for a given $j$ to obtain
$S^{j} = \left[
\begin{array}{cc}
  Z^{j}_{ecg} \\
  Z^{j}_{eda}
\end{array}\right]$, 
where $S^{j} \in \mathbb{R}^{n \times m \times d} $, $d$ is the number of modalities, and $n$ and $m$ are the dimensions of $Z^{j}_{ecg}$ (or $Z^{j}_{eda}$). The tensor $S^j$ is subsequently fed to an attention layer where the weights for each modality are computed as \cite{poria2017multi}:
\begin{align}
  &U^j = ReLU(S^j W^j), \\
  &\theta^j = softmax((U^j)^T w^j_u),
  % L_{ndm} &= exp[U^j_{ndm} w^j_u] \\
  % \theta^j &= \frac{L_{ndm}}{\sum_{k=1}^{d} L_{nkm}}
\end{align}
where $W^j \in \mathbb{R}^{d \times d}$ is the learned weight matrix, ReLU is a rectified linear unit activation function, $U^j$ is the projection of the representation set $S^j$, $w^j_u$ $\in \mathbb{R}^{m}$ is a learned weight vector, $softmax()$ computes the softmax along the second axis, and $\theta^j \in \mathbb{R}^{n \times d \times m}$ is the final calculated attention weight tensor. The transpose operation is done along the second and third axes. We set $d=2$ as we have two modalities (ECG and EDA). We split $\theta^j$ into two tensors, $\theta_{ecg} = \theta_{*, 1, *} \in \mathbb{R}^{n\times m}$ and $\theta_{eda} = \theta_{*, 2, *} \in \mathbb{R}^{n \times m}$ which are the attention weights for ECG and EDA, respectively. Accordingly, we define:
\begin{align}
 \label{equ:attention_multiplication}
  \hat{Z}^j_{ecg} &= \theta_{ecg} \odot Z^{j}_{ecg} \\ \nonumber
  \hat{Z}^j_{eda} &= \theta_{eda} \odot Z^{j}_{eda} \\ \nonumber
\end{align}
where $\odot$ denotes the element-wise multiplication. Equation \ref{equ:attention_multiplication} indicates that the ECG and EDA representations are weighted by their computed attention matrices, respectively. 

Then the ECG and EDA inputs to the next convolutional layers are constructed as: 
\begin{align} \label{equ:new_ecg}
  X^{j+1}_{ecg} &= Z^{j}_{ecg} \otimes \hat{Z}^j_{eda} \\ 
  X^{j+1}_{eda} &= Z^{j}_{eda} \otimes \hat{Z}^j_{ecg}, \label{equ:new_eda}
\end{align}
respectively. In the above equations, $\otimes$ denotes concatenation operation,
$X^{j+1}_{ecg}$ represents the concatenated representations of $Z^{j}_{ecg}$ and $\hat{Z}^j_{eda}$, and $X^{j+1}_{eda}$ represents the concatenated representations of $Z^{j}_{eda}$ and $\hat{Z}^j_{ecg}$. The $X^{j+1}_{ecg}$ and $X^{j+1}_{eda}$ are batch normalized before feeding into the next layer.

In summary, our proposed attentive cross-modal connection exploits an intermediate set of learned representations from one modality and uses it in the other modality to boost the overall performance by taking advantage of the complementary information in both streams. Furthermore, our approach applies learned attention weights to both modalities to focus on the more salient modality where needed.

To study the effects of combining modalities at different locations within deep neural networks on the performance, we define three different types of attentive cross-modal connections (Type I, II, and III shown in Figure \ref{fig:attx_types}). These connections share intermediate feature representations between the modality streams. For connection Type I, the AttX block only computes the attention weights for ECG, denotes as $\theta_{ecg}$, that are then multiplied with $Z^{j}_{ecg}$ to obtain the weighted intermediate representations of ECG, denoted by $\hat{Z}^j_{ecg}$. These representations are then shared with the representations of EDA denoted by $Z^{j}_{eda}$. Similarly, for Type II, the weighted intermediate representations $\hat{Z}^j_{eda}$ are shared with the representations of ECG denoted by $Z^{j}_{ecg}$ at a particular location in the network. For connection Type III, both $\theta_{ecg}$ and $\theta_{eda}$ are learned by the AttX block. These attention weights are then used to transform $Z^{j}_{ecg}$ and $Z^{j}_{eda}$ to obtain $\hat{Z}^j_{ecg}$ and $\hat{Z}^j_{eda}$, respectively. Further, these weighted intermediate representations are concatenated with $Z^{j}_{eda}$ and $Z^{j}_{ecg}$ using Equations \ref{equ:new_ecg} and \ref{equ:new_eda}. To specify the exact locations in which AttX connections are added in the network, we divide our network into four different stages. Figure \ref{fig:networkarch} shows the AttX connections Type III at the end of stages 1, 2, and 3. The locations and combinations of the proposed AttX connections will be tuned empirically through several experiments, which will be discussed in Section \ref{sec:experiments}. 

\subsection{Network Architecture Details}
\label{sub:netarch}
As mentioned earlier, we require two different deep neural networks $H_{ecg}$ and $H_{eda}$. The proposed network is depicted in Figure \ref{fig:networkarch}. For simplicity, we first leave out the AttX and batch normalization operations from our description and tend to it once $H_{ecg}$ and $H_{eda}$ have been discussed. Here, $H_{ecg}$ is designed as a 1D CNN with a residual squeeze and excitation (SE-ResNet) \cite{hu2018squeeze} structure. At Stage 1, $H_{ecg}$ uses Convolutional Blocks (Conv. Block) consisting of Conv-BatchNorm-ReLU layers. Next, at Stage 2, an SE Conv. Block (Figure \ref{fig:networkarch} (b)) is used followed by two SE ID Blocks (Figure \ref{fig:networkarch} (b) without the purple dotted block). This is followed by another SE Conv. Block, and an SE ID Block at Stage 3. A global average pooling layer follows this to obtain feature representation $Y_{ecg}$. The representation $Y_{ecg}$ is projected onto a latent space to obtain a smaller dimension, $Y'_{ecg}$, using two FC layers with output sizes of 64 followed by ReLU.
For $H_{eda}$, we design a CNN with 4 main blocks (see Figure \ref{fig:networkarch}). Similar to $H_{ecg}$, the first stage is a single Conv. Block. Next, two consecutive Conv. Blocks are used for each of the following two stages. This is followed by two Conv. Blocks, an SE Block (Figure \ref{fig:networkarch} (c)), and a global average pooling layer. The representation $Y_{eda}$ is projected onto a latent space to obtain a smaller dimension, $Y'_{eda}$, using 2 FC layers with output sizes of 64 followed by ReLU. Finally, the learned representations ($Y'_{ecg}$ and $Y'_{eda}$) are merged and fed to a classifier with 2 FC layers followed by a ReLU activation. The size of both FC layers is 128, and a softmax layer is used for classification. Given the $H_{ecg}$ and $H_{eda}$ models described above, the AttX mechanism (Type I, II, or III) proposed earlier can be integrated into different single stages or combinations of stages, followed by a batch normalization operation. Figure \ref{fig:networkarch} (a) depicts a Type III AttX integrated into all three stages simultaneously.

\section{Experiments}
\label{sec:experiments}
\subsection{Dataset and Pre-Processing}
We use the multimodal dataset on WEarable Stress and Affect Detection (WESAD) \cite{schmidt2018introducing}. This dataset has collected physiological signals from 15 subjects using two different devices worn by the subjects, one on the chest and the other on the wrist. For our study, we only consider the ECG and the EDA data sampled at 700 \textit{Hz}. We perform a number of basic pre-processing steps to clean the ECG and EDA modalities. For ECG, a Butterworth bandpass filter is applied with a passband frequency of 5-15 \textit{Hz} \cite{pan32real, ross2020unsupervised}. The raw EDA signal is filtered using a lowpass filter with a cut-off frequency of 3 \textit{Hz}. Both ECG and EDA are normalized using user-specific z-score normalization \cite{sarkar2020self}. ECG and EDA are then re-sampled from 700 \textit{Hz} to 256 \textit{Hz} and are segmented into 10-second windows with 50\% overlap to form individual samples \cite{plataniotis2006ecg}. To create binary stress vs. non-stress classes for evaluation purposes, the amusement and neutral classes were aggregated to create the \textit{non-stress} class.

\begin{table}[!t]
\begin{center} 
\caption{Evaluation of different AttX configurations are presented. The arrows show the direction of shared information.
}
\label{tbl:eval_three_class}
\scriptsize
\begin{tabular}{|c|l|c|c|c|c|c|c|}
\hline
\multicolumn{2}{|c|}{\textbf{Fusion Configuration}} & \multirow{2}{*}{\textbf{Accuracy}} & \multirow{2}{*}{\textbf{Macro F1}} & \multirow{2}{*}{\textbf{Weighted F1}} \\ \cline{1-2}
\textbf{Connection Type} & \textbf{Stage} &  &  & \\ \hline %\hline
Feature-Level Fusion & \multicolumn{1}{c|}{-} & 81.49 & 78.60 & 82.41  
\\ \hline
\multirow{7}{1cm}{Type I (EDA$\rightarrow$ECG)} & 1 & 84.37 & 82.64 & 84.52  
\\ 
% \cline{2-8}
& 2 & 79.67 & 77.95 & 79.72 %& 72.25 & 65.18 & 74.56 
\\ 
% \cline{2-8}
& 3 & 87.63 & 86.39 & 87.90 %& 73.70 & 66.00 & 77.63 
\\
% \cline{2-8}
& 1, 2 & 84.42 & 81.35 & 85.37 %& 72.70 & 64.42 & 76.80 
\\
% \cline{2-8}
& 1, 3 & 84.13 & 81.13 & 85.47 %& 73.94 & 66.66 & 77.38 
\\ 
% \cline{2-8}
& 2, 3 & 88.33 & 86.14 & 89.05 %& 71.73 & 64.35 & 74.45 
\\ 
% \cline{2-8}
& 1, 2, 3 & 88.72 & 87.85 & 88.53 %& 72.84 & 64.85 & 75.83 
\\ \hline
\multirow{7}{1cm}{Type II (ECG$\rightarrow$EDA)} & 1 & 87.85 & 85.90 & 88.05 \\%& 77.23 & 70.54 & 80.12 \\ 
% \cline{2-8}
& 2 & 81.35 & 77.02 & 83.11 %& 74.95 & 67.20 & 78.24 
\\ 
% \cline{2-8}
& 3 & 81.74 & 79.85 & 81.89 %& 70.77 & 62.74 & 74.58 
\\ 
% \cline{2-8}
& 1, 2 & 88.67 & 87.12 & 88.96 %& 77.96 & 72.39 & 80.12 
\\ 
% \cline{2-8}
& 1, 3 & 85.90 & 82.59 & 87.03 % & 76.79 & 69.20 & 79.66 
\\ 
% \cline{2-8}
& 2, 3 & 90.45 & 88.58 & 90.58 %& 74.40 & 66.66 & 77.86 
\\ 
% \cline{2-8}
& 1, 2, 3 & 87.93 & 85.80 & 89.06 %& 78.89 & \textbf{74.52} & 80.61 
\\ \hline
\multirow{7}{1cm}{Type III (ECG$\leftrightarrow$EDA)} & 1 & 88.58 & 86.49 & 88.78 %& 75.62 & 67.87 & 78.62 
\\
% \cline{2-8}
& 2 & 85.42 & 82.14 & 86.09 %& 74.48 & 69.44 & 77.05 
\\ 
% \cline{2-8}
& 3 & 88.12 & 86.78 & 88.12 %& 69.93 & 62.21 & 73.12 
\\ 
% \cline{2-8}
& 1, 2 & \textbf{92.08} & \textbf{91.11} & \textbf{92.08} %& 75.28 & 69.63 & 77.58 
\\ 
% \cline{2-8}
& 1, 3 & 88.53 & 85.94 & 88.78 %& 77.32 & 73.68 & 79.58 
\\ 
% \cline{2-8}
& 2, 3 & 89.06 & 87.24 & 89.12 %& 75.34 & 68.39 & 79.12 
\\ 
% \cline{2-8}
& 1, 2, 3 & 90.00 & 88.66 & 89.91 %& \textbf{79.22} & {72.75} & \textbf{81.96}
\\ \hline %\hline
\end{tabular}
\end{center}
\end{table}

\subsection{Implementation Details and Evaluation}
Our model is implemented using Keras with TensorFlow backend on an NVIDIA GeForce RTX 2080 Ti GPU. To rigorously test our model, we use the Leave-One-Subject-Out (LOSO) evaluation scheme for binary classification. We use a standard Adam optimizer with a learning rate of \num{1e-3} and cross-entropy loss. A batch size of 16 is used, and the network is trained for 100 epochs. For evaluating our method, we use accuracy and F1-score with macro-averaging.

\subsection{Performance and Comparison}
\label{sec:results}
We evaluate the performance of the proposed AttX connections when added to the network. We investigate both the location (stage) at which these connections can be added individually as well as their combinations. First, we add AttX Type I to stages 1 through 3. Next, we experiment by adding two AttX connections simultaneously, to stages 1 \& 2, 1 \& 3, and 2 \& 3. This is followed by adding three AttX connections to stages 1, 2, \& 3 at the same time. Next, we repeat the same experiments for AttX Type II, followed by AttX Type III.
 
Table \ref{tbl:eval_three_class} shows the experimental results of the proposed AttX connections (Type I, II, and III) integrated into different stages of the model. From Table \ref{tbl:eval_three_class}, we can conclude that in most cases sharing information in both directions among modalities helps the network to perform better than sharing in just one direction. In line with the above, we can observe that the best performance for binary classification is obtained by connecting AttX Type III connections to stages 1 \& 2.

Next, we compare our best performing configuration with state-of-the-art methods in Table \ref{tbl:binary_results}. From Table \ref{tbl:binary_results}, we observe that our model achieves an accuracy and F1 score of $92.08\%$ and $91.11$, respectively, which are comparable to \cite{schmidt2018introducing} and outperform \cite{bota2020emotion, bajpai2020evaluating}, despite having used considerably fewer modalities. This points to the fact that our framework has been more effective in capturing the information in the ECG and EDA modalities. 

Lastly, we perform detailed ablations to evaluate the impact of each component in our multimodal solution and present the results in Table \ref{tbl:ablation}. Further, we observe that multi-stage information sharing (ours w/o attention) performs better than the feature-fusion and the uni-modal schemes. Finally, we also present the results of our full AttX with the best configuration, which shows considerable improvement in the results.

\setlength{\tabcolsep}{3pt}
\begin{table}[!t]
\begin{center}
\caption{Stress classification results in comparison to related work using LOSO validation.}
\label{tbl:binary_results}
\scriptsize
\begin{tabular}{|c|c|c|c|}
\hline
\textbf{Method} & \textbf{Modality} & \textbf{Acc.} & \textbf{Mac. F1}\\
\hline %\hline
Bota et al. \cite{bota2020emotion} (feat.) &  EDA, ECG, BVP, RSP & 85.80 & -- \\
Bota et al. \cite{bota2020emotion} (dec.) &  EDA, ECG, BVP, RSP &   87.60 & 19.40 \\
Bajpai et al. \cite{bajpai2020evaluating} & - & 90.00 & 90.00 \\
Schmidt et al. \cite{schmidt2018introducing} (feat.) & ECG, EDA, EMG, RESP, ACC, T & 93.12 & 91.47 \\ 
Ours (Type III, stages 1,2) &	EDA, ECG & 92.08	& 91.11 \\
\hline %\hline
\end{tabular}
\end{center}
\end{table}

\setlength{\tabcolsep}{3pt}
\begin{table}[!t]
\begin{center}
\caption{Ablation experiments on our proposed pipeline.}
\label{tbl:ablation}
\begin{tabular}{|c|c|c|c|}
\hline
\textbf{Method} & \textbf{Modality} 
& \textbf{Accuracy} & \textbf{Mac. F1} 
\\
\hline 
Unimodal & EDA & 78.65 & 70.47  
\\
Unimodal & ECG & 78.51 & 76.96  
\\
Feat. Fus. &	EDA, ECG & 81.49 & 78.60 
\\
Ours w/o att. &	EDA, ECG &	87.09 &	83.10 
\\
Ours (AttX) &	EDA, ECG & 92.08 & 91.11 
\\ \hline
\end{tabular}
\end{center}
\end{table}

\section{Conclusion and Future Work}
This paper proposed an attentive cross-modal connection, AttX, to learn strong shared representations from ECG and EDA for the multimodal classification of emotions. We proposed a model consisting of individual pipelines for ECG and EDA, and employed AttX connections to regulate the flow of intermediate information between the two pipelines. Three different variations of AttX were evaluated (Type I, II, and III). Our analysis suggests that the best performance is obtained when multiple AttX connections are added at different locations in the network. Moreover, AttX Type III proved more effective than Type I and II, indicating that bi-directional information sharing is more beneficial to the model as a whole. Lastly, in comparison to other multimodal solutions with even more modalities, our method showed comparable or better results.

It should be noted that while we observed interesting trends in the type, location, and number of AttX connections required to obtain optimum performance, additional studies can be performed in future work to further investigate its generalizability and impact on other datasets and modalities. 

\section*{Acknowledgment}
We thank Innovation for Defence Excellence and Security (IDEaS) program for funding this project.

\bibliographystyle{IEEEtran}
\small{
\bibliography{references}
}

\end{document}